\documentclass{article} 
\usepackage{proceedings}
\usepackage{latexsym}

\newtheorem{definition}{Definition}
\newtheorem{proposition} {Proposition}

\newcommand{\setslash}{\mbox{ $|$ }}

\newcommand{\QCL}{{\it QCL}}

\newcommand{\LPOD}{{\it LPOD}}
\newcommand{\Smodels}{\emph{Smodels}}

\newcommand{\no}{{\rm not\,}}

\newenvironment{proof}{{\bf Proof\mbox{:\hspace{.02in}} }}{\( \Box \) 
\vspace{.01in}}

\title{Logic Programming with Ordered Disjunction}
\author{Gerhard Brewka\thanks{This is a revised and extended version of a paper presented at AAAI-02. The paper was also presented at NMR-02, the Intl. Workshop on Nonmonotonic Reasoning} \\
Universit\"at Leipzig \\ 
Institut f\"ur Informatik \\
Augustusplatz 10-11 \\
04109 Leipzig, Germany \\
brewka@informatik.uni-leipzig.de}

\begin{document}

\maketitle

\begin{abstract}
Logic programs with ordered disjunction ($\LPOD$s) combine ideas underlying 
Qualitative Choice Logic \cite{BreBenBer02} and answer set programming.  Logic 
programming under answer set semantics is extended with a new connective 
called ordered disjunction. The new connective allows us to represent 
alternative, ranked 
options for problem solutions in the heads of rules: $A \times B$ intuitively 
means: if 
possible $A$, but if $A$ is not possible then at least $B$. The semantics of 
logic programs with ordered disjunction is based on a preference relation on 
answer sets. $\LPOD$s are useful for applications in design and configuration 
and can serve as a basis for qualitative decision making.
\end{abstract}

\section{Introduction}
In a recent paper \cite{BreBenBer02} a propositional logic called Qualitative 
Choice Logic ($\QCL$) was introduced. The logic contains a new connective 
$\times$ representing ordered disjunction. Intuitively, $A \times B$ stands 
for: if possible $A$, but if $A$ is impossible then (at least) $B$. This 
connective allows context dependent preferences to be represented in a simple 
and elegant fashion. As a simple example consider the preferences for booking 
a hotel for a conference. Assume the most preferred option is to be within 
walking distance from the conference site, the second best option is to have 
transportation provided by the hotel, the third best is public transportation. 
This can simply be represented as

\[walking \times hotel\mbox{$-$}transport \times public\mbox{$-$}transport \]
From a description of available hotels, a disjunction expressing that one of 
the hotels must be picked,  and the above formula $\QCL$ is able to derive the 
hotel which satisfies best the given preferences (if there is more than one 
such hotel a corresponding disjunction is concluded).

The semantics of the logic is based on degrees of satisfaction of a formula in 
a classical model. The degrees, intuitively, measure disappointment and induce 
a preference relation on models. Consequence is defined in terms of most 
preferred models. It is argued in that paper that there are numerous useful 
applications, e.g. in configuration and design.

In this paper we want to combine ideas underlying $\QCL$ with logic 
programming. More precisely, we want to investigate logic programs based on 
rules with ordered disjunction in the heads. We call such programs logic 
programs with ordered disjunction ($\LPOD$s).

The semantical framework in which the investigation will be carried out is 
that of answer set semantics \cite{GelLif91}. Logic programs under answer set 
semantics have emerged as a new promising programming paradigm dubbed answer 
set programming. There are numerous interesting AI applications of answer set 
programming, for instance in planning \cite{Lif01} and configuration 
\cite{Soi00}. One of the reasons for this success is the availability of 
highly efficient  systems for computing answer sets like {\em smodels} 
\cite{NieSim97} and {\em dlv} \cite{Eiteretal98}. 


We think it is worthwhile to investigate simple representations of context 
dependent preferences in the answer set programming paradigm.
Our combination of ideas from $\QCL$ and answer set programming will lead to 
an approach which is less expressive than $\QCL$ in one respect: the syntax of 
$\LPOD$s restricts the appearance of ordered disjunction to the head of rules. 
On the other hand, we inherit from answer set programming the nonmonotonic 
aspects which are due to default negation. This allows us to combine default 
knowledge with knowledge about preferences and desires in a simple and elegant 
way.

The basic intuition underlying our approach can be described as follows: we 
will use the ordered disjunctions in rule heads to select some of the answer 
sets of a program as the preferred ones. Consider a program containing the 
rule \[A \times B \leftarrow C\] If $S_1$  is an answer set containing $C$ 
and $A$ and $S_2$ is an answer set containing $C$ and $B$ but not $A$, 
then - ceteris paribus (other things being equal) - $S_1$ is preferred 
over $S_2$. Of course, we have to give precise meaning to the ceteris 
paribus phrase. Intuitively ceteris paribus is to be read as $S_1$ and 
$S_2$ satisfy the other rules in the program equally well.

We will show that under certain conditions reasoning from most preferred 
answer sets yields optimal problem solutions. In more general decision making 
settings the preference relation on answer sets provides a basis for best 
possible choices given a specific decision strategy.

We will restrict our discussion in this paper to propositional programs. 
However, as usual in answer set programming, we admit rule schemata containing 
variables bearing in mind that these schemata are just convenient 
representations for the set of their ground instances.

The rest of the paper is organized as follows. In the next section we 
introduce syntax and semantics of $\LPOD$s. We define the degree of 
satisfaction of a rule in an answer set and show how to use the degrees to 
determine a preference relation on answer sets. Conclusions are defined as the 
literals true in all preferred answer sets. The subsequent section discusses 
some simple examples and potential applications. We then investigate implementation issues. The following section 
shows how $\LPOD$s can serve as a basis for a qualitative 
decision making. The last section discusses related work and concludes.

\section{Logic programs with ordered disjunction} 
 
Logic programming with ordered disjunction is an extension of logic 
programming with two kinds of negation (default and strong negation) 
\cite{GelLif91}. The new connective $\times$ representing ordered disjunction 
is allowed to appear in the head of rules only. A (propositional) $\LPOD$ thus 
consists of rules of the form

\[C_1 \times \ldots \times C_n \leftarrow A_1,\ldots, A_m, \no B_1, \ldots, 
\no B_k\]
where the $C_i$, $A_j$ and $B_l$ are ground literals.

The intuitive reading of the rule head is: if possible $C_1$, if $C_1$ is not 
possible then $C_2$, ..., if all of $C_1, \ldots, C_{n-1}$ are not possible 
then $C_n$. The literals $C_i$ are called choices of the rule. Extended logic 
programs with two negations are a special case where $n=1$ for all rules. As 
usual we omit $\leftarrow$ whenever $m = 0$ and $k = 0$, that is, if the rule 
is a fact. Moreover, rules of the form $\leftarrow body$ (constraints) are 
used as abbreviations for $p \leftarrow body,  \no p$ for some $p$ not 
appearing in the rest of the program. The effect is that no answer sets 
containing $body$ exist.

Before defining the semantics of $\LPOD$s a few observations are in order. As 
already mentioned in the introduction we want to use the ranking of literals 
in the head of rules to select some of the answer sets of a program as the 
preferred ones. But what are the answer sets of a program among which to make 
this selection?

Since ordered disjunction is a particular prioritized form of disjunction it 
seems like a natural idea to base the semantics of $\LPOD$s on one of the 
standard semantics for disjunctive logic programs, for instance Gelfond and 
Lifschitz's semantics \cite{GelLif91}. 

Unfortunately, this doesn't work. The problem is that most of the semantics 
for disjunctive logic programs have minimality built in. For instance, 
according to Gelfond and Lifschitz, $S$ is an answer set of a disjunctive 
logic program $P$ iff $S$ is a minimal set of literals which is  logically 
closed, and closed under the $S$-reduct of $P$.
The $S$-reduct of $P$ is obtained from $P$ by 
(1) deleting all rules $r$ from $P$ such that $\no B_j$ in the body of $r$ and 
$B_j \in S$, and (2) deleting all default negated literals from the remaining 
rules.
A set of literals $S$ is closed under a rule $r$ if one of the literals in the 
head of $r$ is in $S$ whenever the body is true in $S$ (see \cite{GelLif91} 
for the details). 

In this approach answer sets are minimal: if $S_1$ and $S_2$ are answer sets 
of a disjunctive program $P$ and $S_1 \subseteq S_2$, then $S_2 \subseteq 
S_1$.

Minimality is not always wanted for $\LPOD$s. Consider the following two 
facts:

\begin{quote}
1)      $A \times B \times C$ \\
2)      $B \times D$
\end{quote}
The single best way of satisfying both ordered disjunctions is obviously to 
make $A$ and $B$ true, that is, we would expect $\{A,B\}$ to be the single 
preferred answer set of this simple $\LPOD$. However, since $B$ is sufficient 
to satisfy both disjunctions, the set $\{A,B\}$ is not even an answer set of  
the corresponding disjunctive logic program (where $\times$ is replaced by 
$\vee$) according to the semantics of \cite{GelLif91}: the built in minimality 
precludes sets containing both $A$ and $B$ from consideration.

We thus have to use a semantics which is not minimal. Indeed, there is such a 
semantics, the possible models semantics proposed by Sakama and Inoue 
\cite{SakIno94}. It is based on so-called split programs, that is, disjunction 
free programs which contain arbitrary subsets of single head rules obtained 
from disjunctive rules by deleting all but one alternatives in the head. 

Unfortunately, also this semantics is inadequate, this time for opposite 
reasons: it admits too many literals in answer sets. Consider the disjunctive 
logic program
\begin{quote}
1)      $A \vee B \vee C$
\end{quote}
There are seven split programs corresponding to the nonempty subsets of the 
literals of the fact. The split program containing the facts $A, B, C$ 
generates the possible model where $A,B,C$ is true. 

Let us replace disjunction by ordered disjunction in this formula. According 
to our intuitive discussion we want to read the rule as "if possible $A$, if 
this is not possible then $B$, and if also $B$ is not possible then $C$". 
Under this reading models containing more than one of the literals in the head 
do not seem justified on the basis of a single rule (they may be justified by 
different rules, though).

For this reason we will not allow cases where a single rule of the original 
program gives rise to more than one rule in the split program. There is a 
further complication: consider the program:
\begin{quote}
1)   $A \times B \times C$ \\
2) $A$
\end{quote}
We do not want to obtain $\{A,B\}$ as an answer set from the split program 
consisting of these 2 atomic facts since again this does not correspond to the 
intuitive reading of the first rule ($B$ only if $A$ is not possible). We 
therefore have to use slightly more complicated rules in split programs.
\begin{definition}
Let  $r = C_1 \times \ldots \times C_n \leftarrow body$ be a rule. For $k \leq 
n$ we define the $k$th option of $r$ as \[r^k = C_k \leftarrow body, \no C_1, 
\ldots, \no C_{k-1}.\]
\end{definition}
\begin{definition}
Let $P$ be an $\LPOD$. $P'$ is a split program of $P$ if it is obtained from 
$P$ by replacing each rule in $P$ by one of its options.
\end{definition}
Here is a simple example. Let $P$ consist of the rules
\begin{quote}
1)      $A \times B \leftarrow \no C$ \\
2)      $B \times C \leftarrow \no D$
\end{quote}
We obtain 4 split programs
\begin{tabbing}
xxx \= xxxxxxxxxxxxxxxxxxxx \= xxxxxxxxxxxxxx\= xxxxxxxxxxxxx\= \kill
\> $A \leftarrow \no C$ \> $A \leftarrow \no C$  \\
\> $B \leftarrow \no D$ \> $C \leftarrow \no D, \no B$ \\
\>   $\mbox{ }$ \\
 \> $B \leftarrow \no C, \no A$  \> $B \leftarrow \no C, \no A$ \\
\> $B \leftarrow \no D$ \> $C \leftarrow \no D, \no B$
\end{tabbing} 

Split programs do not contain ordered disjunction. We thus can define:
\begin{definition}
Let $P$ be an $\LPOD$. A set of literals $A$ is an answer set of $P$ if it is 
a consistent answer set of a split program $P'$ of $P$.
\end{definition}
We exclude inconsistent answer sets from consideration since they do not 
represent possible problem solutions. In the example above we obtain 3 answer 
sets: $\{A,B\},\{C\},\{B\}$. Note that one of the answer sets is a proper 
subset of another answer set. On the other hand, none of the rules in the 
original $\LPOD$ sanctions more than one literal in any of the answer sets, as 
intended.

Not all of the answer sets satisfy our most intended options. Clearly, 
$\{B,A\}$ gives us the best options for both rules, whereas $\{C\}$ gives only 
the second best option for 2) and $\{B\}$ the second best option for 1). To 
distinguish between more and less intended answer sets we introduce the degree 
of satisfaction of a rule in an answer set:

\begin{definition}
Let $S$ be an answer set of an $\LPOD$ $P$. $S$ satisfies the rule
\[C_1 \times \ldots \times C_n \leftarrow A_1, \ldots, A_m, \no B_1, \ldots, 
\no B_k\]
\begin{itemize}
\item to degree 1 if $A_j \not \in S$, for some j, or $B_i \in S$, for some i,
\item to degree j $(1 \leq  j \leq n)$ if all $A_j \in S$, no $B_i \in S$, and 
$j = min\{ r \setslash C_r \in S\}$.
\end{itemize}
\end{definition}

\begin{proposition}
If $A$ is an answer set of $P$ then $A$ satisfies all rules of $P$ to some 
degree.\footnote{The other direction of the proposition does obviously not 
hold. For example, the set $\{A\}$ satisfies the rule $B \leftarrow \no A$, 
but is not an answer set for the program consisting of this single rule.}
\end{proposition}
\begin{proof}
Let $r$ be a rule of $P$. If $S$ is an answer set of $P$, then there is a 
split program $P'$ such that $S$ is an answer set of $P'$. Let $r^i$ be the 
rule in $P'$ generated from $r$. Since $S$ is an answer set of $P'$ either the 
body of $r^i$ is satisfied in $S$ and thus $C_i$ is contained in $S$, in which 
case $r$ is satisfied to degree $i$ or smaller, or the body of $ r^i $ is not 
satisfied in $S$, in which case $r$ is satisfied to degree 1 in $S$, or there 
is a better choice than $C_k$, $k < i$, in $S$ and $r$ is satisfied to degree 
$k$.
\end{proof}
We use the degrees of satisfaction of a rule to define a preference relation 
on answer sets. There are different ways of doing this. For instance, we could 
simply add up the satisfaction degrees of all rules and prefer those answer 
sets where the total sum is minimal. Although this may be reasonable in 
certain applications, this approach makes quite strong assumptions about the 
commensurability of choices in different rule heads. In \cite{BreBenBer02} a 
lexicographic ordering of models based on the number of premises satisfied to 
a particular degree was proposed. This lexicographic ordering has a highly 
syntactic flavour. Therefore, we will use here a somewhat more cautious 
preference relation (in the sense that fewer answer sets are considered better 
than others) based on set inclusion of the rules satisfied to certain degrees:

\begin{definition}
For a set of literals $S$, let $S^i(P)$ denote the set of rules in $P$ satisfied by 
$S$ to degree $i$. Let $S_1$ and $S_2$ be answer sets of an $\LPOD$ $P$.
$S_1$ is preferred to $S_2$ ($S_1 > S_2$) iff there is $i$ such that $S_2^i(P 
) \subset S_1^i(P)$, and for all $j < i$, $S_1^j(P) = S_2^j(P)$.
\end{definition}
\begin{definition}
A set of literals $S$ is a preferred answer set of an $\LPOD$ $P$ iff $S$ is 
an answer set of $P$ and there is no answer set $S'$ of $P$ such that $S' > 
S$.
\end{definition}
\begin{definition}
A literal $l$ is a conclusion of an $\LPOD$ $P$ iff $l$ is contained in all 
preferred answer sets of $P$.
\end{definition}
Consider again the program
\begin{quote}
1)  $ A \times B \leftarrow \no C$ \\
2)  $ B \times C \leftarrow \no D$
\end{quote}
As discussed before we obtain the 3 answer sets: $S_1 = \{A, B\}$, $S_2 = 
\{C\}$ and  $S_3 = \{B\}$. $S_1 $ satisfies both rules with degree 1, $\{C\}$ 
satisfies 1) to degree 1 but 2) to degree 2. $\{B\}$ satisfies 1) to degree 2 
and 2) to degree 1. The single preferred answer set is thus $S_1$, as 
intended, and $A$ and $B$ are the conclusions of the program.

\section{Examples} \label{sec:examples}

$\LPOD$s allow us - like normal logic programs - to express incomplete and 
defeasible knowledge through the use of default negation. In addition, they 
provide means to represent preferences among intended properties of problem 
solutions. Moreover, these preferences may depend on the current context.

In this section we discuss several examples illustrating potential uses of 
$\LPOD$s. The first example is about how to spend a free afternoon. You like 
to go to the beach, but also to the cinema. Normally you prefer the cinema 
over the beach, unless it is hot (which is the exception in the area where you 
live, except during the summer). If it is hot the beach is preferred over the 
cinema. In summer it is normally hot, but there are exceptions. If it rains 
the beach is out of question. This information can be represented using the 
following rules:

\begin{quote}
1) $cinema \times beach \leftarrow \no hot$ \\
2) $beach \times cinema \leftarrow hot$ \\
3) $hot \leftarrow \no \neg hot, summer$ \\
4) $\neg beach \leftarrow rain$ 
\end{quote}
Without further information about the weather we obtain the single preferred 
answer set $S_1 = \{cinema\}$. There is no information that it might be hot, 
so rule 1) will determine the preferences. $S_1$ satisfies all rules to degree 
1.

Now assume the fact $summer$ is additionally given. In this case we obtain $ 
S_2 = \{summer, hot, beach\}$ as the single preferred answer set. Again this 
answer set satisfies all rules to degree 1.

Next assume that, in addition to $summer$ also the literal $\neg hot$ is 
given. The single preferred answer set now is $S_3 = \{summer, \neg hot, 
cinema\}$. All rules are staisfied to degree 1.

Finally, assume the additional facts are $summer$ and $rain$. Now the single 
preferred answer set (and in fact the single answer set) is \[S_4 = \{summer, 
rain, hot, \neg beach, cinema\}.\] Note that this time it is not possible to 
satisfy all rules to degree 1: rule 2) is satisfied to degree 2 only. As often 
in real life, there are situations where the best options simply do not work 
out.
 
We think that $\LPOD$s are very well suited for representing problems where a 
certain choice has to be made or, more generally, where a number of components 
have to be chosen for a certain configuration task. The general idea would be 
to have

\begin{itemize}

\item for each component a set of rules describing its properties, 

\item rules describing which components are needed for the configuration to be 
complete; this may depend on other components chosen, 

\item rules describing intended properties of the solution we want to 
generate. The involved preferences may be context dependent, and

\item a description of the case at hand.

\end{itemize}
In each case default knowledge can be used to describe what is normally the 
case. Consider the problem of configuring a menu. The menu should consist of a 
$starter$, a $main$ course, a $dessert$ and a $beverage$. As a $starter$ you 
prefer $soup$ over $salad$. As $main$ course $fish$, $beef$ and $lasagne$ are 
possible (this is all you are able to cook) and your preferences are in this 
order. Of course, if the visitor is $vegetarian$ the first two (as well as the 
$soup$) are out of the question. In case of $beef$ you prefer $red$ wine over 
$white$ wine over mineral $water$, otherwise the order between wines is 
reversed. Only $ice\mbox{$-$}coffee$ and $tiramisu$ is available as a 
$dessert$. If $tiramisu$ is chosen, then an extra $coffee$ is necessary. You 
prefer $espresso$ over $cappucino$.

The possible components thus are $soup$, $salad$, $fish$, $beef$, $lasagne$, 
$ice\mbox{$-$}coffee$, $tiramisu$, $espresso$, $cappucino$, $red$, $white$ and 
$water$. The following properties of the components are relevant:
\begin{tabbing}
xx \= xxxxxxxxxxxxxxxxxxxx \= xxxxxxxxxxxxxx\= xxxxxxxxxxxxx\= \kill
\> $\neg vegetarian \leftarrow beef$ \> $alcohol \leftarrow white$  \\
\> $\neg vegetarian \leftarrow fish$ \> $alcohol \leftarrow red$ \\
\>   $\neg vegetarian \leftarrow soup$ 
\end{tabbing} 
The needed components are
\begin{tabbing}
xx \= xxxxxxxxxxxxxxxxxxxx \= xxxxxxxxxxxxxx\= xxxxxxxxxxxxx\= \kill
\> $starter$ \> $beverage$  \\
\> $main$ \> $coffee \leftarrow tiramisu$ \\
\>   $dessert$ 
\end{tabbing} 
The preferences are as follows:
\begin{tabbing}
xx \= xxxxxxxxxxxxxxxxxxxx \= xxxxxxxxxxxxxx\= xxxxxxxxxxxxx\= \kill
\> $soup \times salad \leftarrow starter$ \\
\> $fish \times beef \times lasagne \leftarrow main$ \\
\> $red \times white \times water \leftarrow beverage, beef$ \\
\> $white \times red \times water \leftarrow beverage, \no beef$ \\
\> $espresso \times cappuccino \leftarrow coffee$ \\
\> $ice\mbox{$-$}coffee \leftarrow \no tiramisu, dessert$ \\
\> $tiramisu \leftarrow \no ice\mbox{$-$}coffee, dessert$
\end{tabbing}
Now, given a description of the case at hand, e.g. whether the visitor is 
vegetarian or not, drinks alcohol or not, likes fish etc. the preferred answer 
sets will determine a menu which satisfies the preferences as much as 
possible.
The last two rules are necessary to make sure that one of the desserts is 
picked. For the other courses this is implicit in the specified preferences. 
In the language of \cite{NieSim00} these rules can be represented as the 
cardinality constraint rule $1 \{ice\mbox{$-$}coffee,tiramisu\} 1 \leftarrow 
dessert$. Combinations of $\LPOD$s and such constraints are a topic of further 
research.

\section{Computation} \label{sec:implementation}

The first question to ask is whether $\LPOD$s can simply be reduced to 
standard logic programs with two kinds of negation. In that case standard 
answer set programming techniques would be sufficient for computing 
consequences of $\LPOD$s. We will show that a seemingly natural translation 
does not yield the intended answer sets.

\begin{definition}
The pseudo-translation $trans(r)$ of a rule \[r = C_1 \times \ldots \times C_n 
\leftarrow body\] is the collection of rules
\begin{tabbing}
xxx\=xxxxx\=xxxxxxxx\= \kill
\>$C_1$  \>$\leftarrow  body, \no -C_1$ \\
\>$C_2 $  \>$ \leftarrow  body, \no -C_2,-C_1$ \\
\>$ \ldots$ \\
\>$C_{n-1}$  \>$ \leftarrow  body, \no -C_{n-1},-C_1, \ldots ,-C_{n-2} $ \\
\>$C_n $  \>$\leftarrow  body, -C_1, \ldots,-C_{n-1}$
\end{tabbing}
where $-C$ is the complement of $C$, that is $\neg C$ if $C$ is an atom and 
$C'$ if $C = \neg C'$. The pseudo-translation $trans(P)$ of an $\LPOD$ $P$ is
\[trans(P) = \bigcup_{r \in P} trans(r)\]
\end{definition}
The pseudo-translation creates for each option $C_i$ in the head of $r$  a 
rule with head $C_i$ which has the negation of the better options as 
additional body literals. In addition, the rule is made defeasible by adding 
the default negation of the complement of $C_i$ to the body. There is an 
exception: the rule generated for the last option is not made defeasible this 
way since at least one of the options must be true whenever the body of the 
original rule is true.

Although this translation seems natural it does not work. Consider the 
following example:

\begin{quote}
1) $a \times b$ \\
2) $p \leftarrow \no p, a$
\end{quote}
The single preferred answer set is $\{b\}$. The pseudo-translation is
\begin{quote}
1) $a \leftarrow \no \neg a$ \\
2) $b \leftarrow \neg a$ \\
3) $p \leftarrow \no p, a$
\end{quote}
The resulting program has no answer set. In fact, we can prove the following proposition:
\begin{proposition}
There is no translation $trans$ from $\LPOD$s to extended logic programs (without ordered disjunction) such that for each program $P$ the preferred answer sets of $P$ and the answer sets of $trans(P)$ coincide.
\end{proposition}
\begin{proof}
The proposition follows from the fact that preferred answer sets of $\LPOD$s are not necessarily subset minimal. Consider the program $a \times b$; $c \times b \leftarrow a$; $\neg c$. The preferred answer sets are $S_1 = \{b, \neg c\}$ and $S_2 = \{a, b, \neg c\}$. Clearly, $S_1 \subset S_2$. There is thus no extended logic program with these answer sets.
\end{proof}

Of course, this does not exclude the possibility of translations to programs containing some extra atoms. This is a topic of further study.

An implementation of $\LPOD$s on top of a standard answer set prover for non-disjunctive programs is described in \cite{BreNieSyr02}. 
We compute preferred answer sets of an $\LPOD$ $P$ using two programs.  A similar approach is used
in \cite{JNSY00} to compute stable models of disjunctive logic
programs using \emph{Smodels}. The two programs are:
\begin{itemize}
\item A \emph{generator} $G(P)$ that creates all answer sets of $P$; and
\item A \emph{tester} $T(P,M)$ that checks whether a given answer set
  $M$ of $P$ is maximally preferred.
\end{itemize}

The two programs are run in an interleaved fashion. First, the
generator constructs an arbitrary answer set $M$ of $P$. Next, the tester tries
to find an answer set $M'$ that is strictly better than $M$. The tester possesses an answer set $M'$ iff $M'$ is an answer set of $P$ preferred to $M$. If there is no
such $M'$, we thus know that $M$ is a preferred answer set. Otherwise, we
use $G(P)$ to construct the next candidate. When we want to find only
one preferred answer set we can save some effort by taking $M'$
directly as the new answer set candidate. We can thus iterate until a maximally preferred answer set
is reached. 

Since the tester is based on a declarative representation
of the preference criterion it is easy to switch between different
notions of preference, or to define new ones.

We have constructed a prototype implementation for $\LPOD$s based on
\Smodels, an efficient ASP solver developed at Helsinki University of
Technology. The generator and tester programs use special rule types
of the \Smodels\ system, but they can be modified to work with any ASP
solver. The prototype implementation is available at
http://www.tcs.hut.fi/Software/smodels/priority. The mentioned paper contains also complexity results related to $\LPOD$s.

\section{Decision Making using LPODs} \label{sec:decision}
In Section \ref{sec:examples} we discussed several examples illustrating the 
notions underlying LPODs. The examples were chosen in such a way that the most 
preferred answer sets in each case provided the best solutions to the problem 
at hand. Later in this section we will analyze why this worked for the chosen 
examples.

In more general decision making settings it is not sufficient to consider the 
most preferred answer sets only since this amounts to an extremely optimistic 
view about how the world will behave (this view is sometimes called wishful 
thinking). As is well-known in decision theory, for realistic models of 
decision making it is necessary to clearly distinguish what is under the 
control of the agent (and thus may constitute the agent's decision) from what 
is not. We will do this by distinguishing a subset of the literals in a 
program as decision literals.

In this section we describe a general methodology for qualitative decision 
making based on $\LPOD$s. The basic idea is to use $\LPOD$s to describe 
possible actions or decisions and their consequences, states of the world and 
desired outcomes. The representation of desires induces, through ordered 
disjunction, a preference ordering on answer sets representing their 
desirability. Based on this preference ordering an ordering on possible 
decisions can be defined based on some decision strategy.

Let us describe the necessary steps more precisely:
\begin{enumerate}
\item Among the literals in the logical language distinguish a set of decision 
literals $C$.  $C$ is the set of literals the agent can decide upon. It's the 
agent's decision which makes them true. A decision is a consistent subset of 
$C$. 
\item Represent the different alternative decisions which can be made by the 
agent. This can be done using standard answer set programming techniques. Note 
that certain options may lead to additional choices that need to be made.
\item Represent the different alternative states of the world. Again standard 
answer set programming techniques apply.
\item Represent relationships between and consequences of different 
alternatives.
\item Represent desired properties. This is where ordered disjunction comes 
into play. Of course, desires may be context-dependent.
\item Use the preference relation on answer sets derived form the satisfaction 
degrees of rules to induce a preference relation on possible decisions. Of 
course, there are different ways to do this corresponding to different 
attitudes of the agent towards risk. 
\item  Pick one of the most preferred decisions. 
\end{enumerate}
We will use Savage's famous rotten egg example \cite{Sav54} to illustrate this 
methodology. An agent is preparing an omelette. 5 fresh eggs are already in 
the omelette. There is one more egg. It is uncertain whether this egg is fresh 
or rotten. The agent can 
\begin{itemize}
\item add it to the omelette which means the whole omelette may be wasted, or 
\item throw it away, which means one egg may be wasted, or 
\item put it in a cup, check whether it is ok or not and put it to the 
omelette in the former case, throw it away in the latter. In any case, a cup 
has to be washed if this option is chosen.
\end{itemize}
In this example, the decision literals correspond to the three possible 
actions, that is $C$ is the set of literals built from 
$\{in\mbox{$-$}omelette, in\mbox{$-$}cup, throw\mbox{$-$}away\}$. Here are the 
rules which generate the possible decisions and states of the world:
\begin{tabbing}
xx \= xxxxxxxxxxxxxxxxxxxxx \= xxxxxxxxxxxxxx\= xxxxxxxxxxxxx\= \kill
\> $in\mbox{$-$}omelette \leftarrow \no in\mbox{$-$}cup, \no 
throw\mbox{$-$}away$ \\
\> $in\mbox{$-$}cup \leftarrow \no in\mbox{$-$}omelette, \no 
throw\mbox{$-$}away$ \\
\> $throw\mbox{$-$}away \leftarrow \no in\mbox{$-$}cup, \no 
in\mbox{$-$}omelette$ \\
\> $rotten \leftarrow \no fresh$ \\
\> $fresh \leftarrow \no rotten $
\end{tabbing}
For our example it is not necessary to specify that the different actions and 
states of the egg are mutually exclusive. It is guaranteed by the rules that 
only one of the exclusive options is contained in an answer set. 

We next 
define the effects of the different choices:
\begin{tabbing}
xx \= xxxxxxxxxxxxxxxxxxxxx \= xxxxxxxxxxxxxx\= xxxxxxxxxxxxx\= \kill
\> $5\mbox{$-$}omelette \leftarrow throw\mbox{$-$}away$ \\
\> $6\mbox{$-$}omelette \leftarrow fresh, in\mbox{$-$}omelette$ \\
\> $0\mbox{$-$}omelette \leftarrow rotten, in\mbox{$-$}omelette$ \\
\> $6\mbox{$-$}omelette \leftarrow fresh, in\mbox{$-$}cup$ \\
\> $5\mbox{$-$}omelette \leftarrow rotten, in\mbox{$-$}cup$ \\
\> $\neg wash \leftarrow \no in\mbox{$-$}cup$ \\
\> $wash \leftarrow in\mbox{$-$}cup$
\end{tabbing}
For the different omelettes we must state that they are mutually inconsistent. 
We omit the 6 rules necessary for representing this. They are of the form 
$\neg x\mbox{$-$}omelette \leftarrow y\mbox{$-$}omelette$ with $x \neq y$. We 
finally represent our desires:
\begin{tabbing}
xx \= xxxxxxxxxxxxxxxxxxxxx \= xxxxxxxxxxxxxx\= xxxxxxxxxxxxx\= \kill
\> $ \neg wash \times  wash$ \\
\> $6\mbox{$-$}omelette \times  5\mbox{$-$}omelette \times  
0\mbox{$-$}omelette$
\end{tabbing}
This logic program has the following 6 answer sets
\begin{tabbing}
xx \= xxxxxxxxxxxxxxxxxxxxx \= xxxxxxxxxxxxxx\= xxxxxxxxxxxxx\= \kill
\> $S_1 = \{6\mbox{$-$}omelette, \neg wash, fresh, in\mbox{$-$}omelette \}$ \\
\> $S_2 = \{0\mbox{$-$}omelette, \neg wash, rotten, in\mbox{$-$}omelette \}$ 
\\
\> $S_3 = \{6\mbox{$-$}omelette, wash, fresh, in\mbox{$-$}cup \}$ \\
\> $S_4 = \{5\mbox{$-$}omelette, wash, rotten, in\mbox{$-$}cup \}$ \\
\> $S_5 = \{5\mbox{$-$}omelette, \neg wash, fresh, throw\mbox{$-$}away \}$ \\
\> $S_6 = \{5\mbox{$-$}omelette, \neg wash, rotten, throw\mbox{$-$}away \}$ 
\end{tabbing}
The preference relation among answer sets is as follows:
$S_1$ is the single maximally preferred answer set.  $S_5$ and $S_6$ are 
preferred to $S_2$ and $S_4$ but incomparable to $S_3$. $S_3$ is preferred to 
$S_4$ but incomparable to $S_5$, $S_6$ and $S_2$. $S_2$ and $S_4$ are 
incomparable. Fig. 1 illustrates these relationships:

\setlength{\unitlength}{.12cm}
\begin{picture}(25,25)(-28,0)
\put(0,0){$S_2$}
\put(10,0){$S_4$}
\put(-3,10){$S_5, S_6$}
\put(10,10){$S_3$}
\put(5,20){$S_1$}
\put(1,3){\line(0,1){6}}
\put(11,3){\line(0,1){6}}
\put(1,13){\line(2,3){4}}
\put(11,13){\line(-2,3){4}}
\put(10,3){\line(-1,1){6}}
\end{picture}

\begin{center}
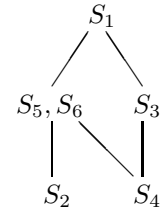
{\em Fig.1: Preferences among answer sets}
\end{center}

Reasoning from maximally preferred answer sets in the example would yield 
$in\mbox{$-$}omelette$ as the alternative chosen by the agent. It is obvious 
that this amounts to an extremely optimistic attitude towards decision making 
which in the example amounts to assuming the egg will be fresh. 

A pessimistic decision maker might choose the action whose worst outcome is 
most tolerable. In the example the answer sets containing 
$throw\mbox{$-$}away$, that is $S_5$ and $S_6$, are preferred to the least 
preferred answer set containing $in\mbox{$-$}omelette$, $S_2$, and to the 
least preferred answer set containing $in\mbox{$-$}cup$, $S_4$. Thus, a 
pessimistic decision maker would choose $throw\mbox{$-$}away$. 

An extremely cautious strategy would prefer a decision $C_1$ over a decision $C_2$ if 
the least preferred answer set(s) containing $C_1$ are preferred to the most 
preferred answer set(s) containing $C_2$. This is a very strong requirement 
and in the egg example no action is preferred to another one according to this 
strategy. 

Finally, we can distinguish a set of state literals $\Sigma$ and compare answer sets statewise (states are subsets of $\Sigma$, the states in the example 
are $fresh$ and $rotten$). A decision $C_1$ is preferred over a decision $C_2$ if for each state $T \subseteq \Sigma$ 
the least preferred answer set(s) containing $C_1 \cup T$ are preferred to the 
most preferred answer set(s) containing $C_2 \cup T$. 

Intuitively, $S_2$ in our example seems far less desirable than $S_4$  and 
both $S_5$ and $S_6$ less desirable than $S_3$. This is not reflected in our 
preference relation on answer sets. To express this it is necessary to 
represent preferences between sets of literals rather than single literals. 
Within our framework this can be done by introducing new atoms representing 
conjunctions of literals. However, it would probably be more elegant to apply 
orderd disjunction directly to sets of literals (read as the conjunction of 
these literals). Extending LPODs in such a way is straightforward.

Another natural idea would be to use numerical penalties. We can use integers 
for this and write, say:
\begin{tabbing}
xx \= xxxxxxxxxxxxxxxxxxxxx \= xxxxxxxxxxxxxx\= xxxxxxxxxxxxx\= \kill
\> $ \neg wash\mbox{$-$}cup \times  wash\mbox{$-$}cup \: (1)$ \\
\> $6\mbox{$-$}omelette \times  5\mbox{$-$}omelette \: (5) \times  
0\mbox{$-$}omelette \: (50)$
\end{tabbing}
The overall penalty for an answer set $S$ is obtained by adding up the 
penalties for all rules, where the penalty of $c_1 \times c_2 (n_2) \times 
\ldots \times c_k (n_k) \leftarrow body$ is 0 if $body$ is not satisfied in 
$S$ or $c_1 \in S$, $n_j$ otherwise, where $j$ is the smallest integer such 
that $c_j \in S$. The preference relation among answer sets is obtained 
through their overall penalty.  In the example we would obtain the following 
overall penalties: 
\begin{tabbing}
xxxxxxxx \= xxxxxxxxx \= xxxxxxxxx\= xxxxxxxx\= \kill
\> $S_1: 0$ \>  $S_3: 1$ \> $S_5: 5$ \\
\> $ S_6: 5$ \> $S_4:6$ \> $S_2 : 50$
\end{tabbing}
Choices could then be ordered on the basis of the average penalties of answer 
sets they contain. This strategy would thus choose $in\mbox{$-$}cup$.

Of course, many alternative strategies can be thought of. A further 
investigation is beyond the scope of this paper and left for future work.

Every approach to qualitative decision making has to combine preferences among 
outcomes of choices with a treatment of uncertainty. In our approach the 
preferences are described through ordered disjunction. But what about the 
uncertainty? Different possible states of the world are represented as 
different answer sets. As usual in nonmonotonic reasoning states of the world 
which are unnormal in some respect are totally disregarded (this is what 
McCarthy called jumping to conclusions). All states which have to be taken 
into account are considered plausible. Further distinctions between 
the generated answer sets are not possible. For instance, it is not possible 
to express, say, that $fresh$ is more probable than $rotten$ in the omelette 
example. If, however the possibility of $rotten$ is negligeable and $fresh$ is 
true by default we can make sure that only answer sets containing $fresh$ are 
generated by using adequate rules. Our general qualitative attitude towards 
uncertainty can thus be described as: states are either negligeable or 
plausible;
in the latter case  no assumption about the degree of plausibility is made.

We are now in a position to analyze why the examples in Sect. 
\ref{sec:examples} which were based on reasoning from most preferred answer 
sets worked out properly. The reason is that in these examples only one answer 
set for the different possible choices (which were left implicit) is 
generated. This means that optimistic, pessimistic and other kinds of LPOD 
based decision making coincide. In general, this is possible whenever there is 
enough knowledge to guarantee a single plausible state for each case at hand 
(as in the cinema example), or whenever all relevant literals are under the 
control of the agent (as in the cooking example).

\section{Conclusion}

In this paper we introduced a new connective to logic programming. This 
connective - called ordered disjunction - can be used to represent context 
dependent preferences in a simple and elegant way. Logic programming with 
ordered disjunction has interesting applications, in particular in design and 
configuration, and it can serve as a basis for qualitative decision models.

There are numerous papers introducing preferences to logic programming. For an 
overview of some of these approaches see the discussion in \cite{BreEit99} or 
the more recent \cite{SchWan01}. Only few of these proposals allow for context 
dependent preferences. Such preferences are discussed for instance in 
\cite{Bre96,BreEit99}. The representation of the preferences in these papers 
is based on 
 the introduction of names for rules, 
 the explicit representation of the preference relation among rules in the 
logical language, and a sophisticated reformulation of the central semantic 
notion (answer set, extension, etc.) with a highly self-referential flavour. 
Alternative approaches \cite{Deletal00,Gro99} are based on compilation 
techniques and make heavy use of meta-predicates in the logical language. 
Nothing like this is necessary in our approach. All we have to do is use the 
degree of satisfaction of a rule to define a preference relation on answer 
sets directly.

Our approach is closely related to work in qualitative decision theory, for an 
overview see \cite{DoyTho99}. Poole \cite{Poo97} aims at a combination of 
logic and decision theory. His approach incorporates quantitative utilities 
whereas our preferences are qualitative. Interestingly, Poole uses a logic 
{\em without} disjunction whereas we {\em enhance} disjunction. In 
\cite{Bouetal99} a graphical representation, somewhat reminiscent of Bayes 
nets,  for conditional preferences among feature values under the {\em ceteris 
paribus} principle is proposed, together with corresponding algorithms. 
$\LPOD$s are more general and offer means to reason defeasibly.
Several models of qualitative decision making based on possibility theory are 
described in \cite{Dubetal99,Benetal00}. They are based on certainty and 
desirability rankings. Some of them make strong commensurability assumptions 
with respect to these rankings. 
In a series of papers \cite{Lan96,TorWey01}, originally
motivated by \cite{Bou94}, the authors propose viewing conditional desires
as constraints on utility functions. Intuitively, $D(a | b)$ stands for: the
$b$-worlds with highest utility satisfy $a$. Our interpretation of ranked
options is very different. Rather than being based on decision theory our
approach can be viewed as giving a particular interpretation to the ceteris
paribus principle. 

In future work we plan to investigate application methodologies for logic 
programming with ordered disjunction.
An answer set programming methodology for configuration tasks has been 
developed in a number of papers by Niemel\"a and colleagues at Helsinki 
University of Technology \cite{Soi00,NieSim00}. We plan to study possibilities 
of combining this methodology with $\LPOD$s. Of course, the discussion of 
qualitative decision models in this paper was very preliminary. We plan to 
work this out in more detail in a separate paper. 

\section*{Acknowledgements}
I would like to thank Salem Benferhat, Richard Booth, Tomi Janhunen, Ilkka Niemel\"{a}, Tommi Syrj\"anen and 
Leon van der Torre for helpful comments.

\bibliography{clp}

\end{document}